\documentclass[10pt, a4paper, conference, compsocconf]{IEEEtran}
\usepackage{cite}
\ifCLASSINFOpdf
\else
\fi

\usepackage{booktabs}
\usepackage{multirow}

\usepackage[normalem]{ulem}

\newcommand\datasetname{\textsc{IndoSum}}

\begin{document}
\title{\datasetname{}: A New Benchmark Dataset for Indonesian Text Summarization}


% \author{\IEEEauthorblockN{Authors Name/s per 1st Affiliation (Author)}
% \IEEEauthorblockA{line 1 (of Affiliation): dept. name of organization\\
% line 2: name of organization, acronyms acceptable\\
% line 3: City, Country\\
% line 4: Email: name@xyz.com}
% \and
% \IEEEauthorblockN{Authors Name/s per 2nd Affiliation (Author)}
% \IEEEauthorblockA{line 1 (of Affiliation): dept. name of organization\\
% line 2: name of organization, acronyms acceptable\\
% line 3: City, Country\\
% line 4: Email: name@xyz.com}
% }

\author{\IEEEauthorblockN{Kemal Kurniawan}
\IEEEauthorblockA{Kata Research Team\\
Kata.ai\\
Jakarta, Indonesia\\
kemal@kata.ai}
\and
\IEEEauthorblockN{Samuel Louvan}
\IEEEauthorblockA{Fondazione Bruno Kessler\\
University of Trento\\
Trento, Italy\\
slouvan@fbk.eu}
}


%Telephone: (800) 555--1212, Fax: (888) 555--1212}


% https://tex.stackexchange.com/questions/279769/ieee-transactions-copyright/325013#325013
\IEEEoverridecommandlockouts
\IEEEpubid{\begin{minipage}{\textwidth}
  \vspace{10mm}
  \copyright~2018 IEEE. Personal use of this material is permitted. Permission
  from IEEE must be obtained for all other uses, in any current or future media,
  including reprinting/republishing this material for advertising or
  promotional purposes, creating new collective works, for resale or
  redistribution to servers or lists, or reuse of any copyrighted component or
  this work in other works. The final version of this article is available at
  https://doi.org/10.1109/IALP.2018.8629109.
\end{minipage}}

\maketitle

\begin{abstract}
  Automatic text summarization is generally considered as a challenging task in
  the NLP community. One of the challenges is the publicly available and large
  dataset that is relatively rare and difficult to construct. The problem is even
  worse for low-resource languages such as Indonesian. In this paper, we present
  \datasetname{}, a new benchmark dataset for Indonesian text summarization. The
  dataset consists of news articles and manually constructed summaries. Notably,
  the dataset is almost 200x larger than the previous Indonesian summarization
  dataset of the same domain. We evaluated various extractive summarization
  approaches and obtained encouraging results which demonstrate the usefulness
  of the dataset and provide baselines for future research. The code and the
  dataset are available online under permissive licenses.
\end{abstract}


% \fi
\IEEEpeerreviewmaketitle

\section{Introduction}
The goal of text summarization task is to produce a summary from a set of
documents. The summary should retain important information and be reasonably
shorter than the original documents~\cite{das2007}. When the set of documents
contains only a single document, the task is usually referred to as
\textit{single-document summarization}. There are two kinds of summarization
characterized by how the summary is produced: \textit{extractive} and
\textit{abstractive}. Extractive summarization attempts to extract few important
sentences verbatim from the original document. In contrast, abstractive
summarization tries to produce an abstract which may contain sentences that do
not exist in or are paraphrased from the original document.

Despite quite a few number of research on Indonesian text summarization, none of
them were trained nor evaluated on a large, publicly available dataset.
Also, although ROUGE~\cite{lin2004} is the standard intrinsic evaluation metric
for English text summarization, for Indonesian it does not seem so. Previous
works rarely state explicitly that their evaluation was performed with ROUGE.
The lack of a benchmark dataset and the different evaluation metrics make
comparing among Indonesian text summarization research difficult.

In this work, we introduce \datasetname{}, a new benchmark dataset for
Indonesian text summarization, and evaluated several well-known extractive
single-document summarization methods on the dataset. The dataset consists of
online news articles and has almost 200 times more documents than the next
largest one of the same domain~\cite{najibullah2015}. To encourage further
research in this area, we make our dataset publicly available. In short, the
contribution of this work is two-fold:
\begin{enumerate}
\item \datasetname{}, a large dataset for text summarization in Indonesian that is
  compiled from online news articles and publicly available.
\item Evaluation of state-of-the-art extractive summarization methods on the
  dataset using ROUGE as the standard metric for text summarization.
\end{enumerate}
The state-of-the-art result on the dataset, although impressive, is still
significantly lower than the maximum possible ROUGE score. This result suggests
that the dataset is sufficiently challenging to be used as evaluation benchmark
for future research on Indonesian text summarization.


\section{Related work}

Fachrurrozi et al.~\cite{fachrurrozi2013} proposed some scoring methods and used
them with TF-IDF to rank and summarize news articles. Another
work~\cite{silvia2014} used latent Dirichlet allocation coupled with genetic
algorithm to produce summaries for online news articles. Simple methods like
naive Bayes has also been used for Indonesian news
summarization~\cite{najibullah2015}, although for English, naive Bayes has been
used almost two decades earlier~\cite{aone1998}. A more recent
work~\cite{gunawan2017} employed a summarization algorithm called TextTeaser
with some predefined features for news articles as well. Slamet et
al.~\cite{slamet2018} used TF-IDF to convert sentences into vectors, and their
similarities are then computed against another vector obtained from some
keywords. They used these similarity scores to extract important sentences as
the summary. Unfortunately, all these work do not seem to be evaluated using
ROUGE, despite being the standard metric for text summarization research.

An example of Indonesian text summarization research which used ROUGE
is~\cite{massandy2014}. They employed the best method on TAC 2011 competition
for news dataset and achieved ROUGE-2 scores that are close to that of humans.
However, their dataset consists of only 56 articles which is very small, and the
dataset is not available publicly.

An attempt to make a public summarization dataset has been done
in~\cite{koto2016}. They compiled a chat dataset along with its summary, which
has both the extractive and abstractive versions. This work is a good step toward
standardizing summarization research for Indonesian. However, to the best of our
knowledge, for news dataset, there has not been a publicly available dataset,
let alone a standard.

\section{Methodology}

\subsection{\datasetname{}: a new benchmark dataset}

\begin{figure*}[!t]\footnotesize
  \centering
  \begin{tabular}{|@{~}p{0.9\textwidth}@{~}|}
    \hline
    \textbf{\uwave{Suara.com - Cerita sekuel terbaru James Bond bocor}} \\
    \uwave{Menurut sumber yang terlibat dalam produksi film ini, agen rahasia 007 berhenti
    menjadi mata-mata Inggris demi menikah dengan perempuan yang dicintainya.} \\
    "Bond berhenti menjadi agen rahasia karena jatuh cinta dan menikah dengan perempuan
    yang dicintai," tutur seorang sumber yang dekat dengan produksi seperti dikutip laman
    PageSix.com. \\
    Dalam film tersebut, Bond diduga menikahi Madeleine Swann yang diperankan oleh Lea
    Seydoux. \\
    Lea diketahui bermain sebagai gadis Bond di sekuel Spectre pada 2015 silam. \\
    \uwave{Jika benar, ini merupakan satu-satunya sekuel yang bercerita pernikahan James
    Bond sejak 1969.} \\
    \uwave{Sebelumnya, di sekuel On Her Majesty, James Bond menikahi Tracy Draco yang
    diperankan Diana Rigg.} \\
    \uwave{Namun, di film itu Draco terbunuh.} \\
    Plot sekuel film James Bond ke-25 bocor tak lama setelah Daniel Craig mengumumkan bakal
    kembali memerankan tokoh agen 007. \\
    \hline \hline
    Cerita sekuel terbaru James Bond bocor. \\
    Menurut sumber yang terlibat dalam produksi film ini, agen rahasia 007 berhenti menjadi
    mata-mata Inggris demi menikah dengan perempuan yang dicintainya. \\
    Jika benar, ini merupakan satu-satunya sekuel yang bercerita pernikahan James Bond
    sejak 1969. \\
    Sebelumnya, di sekuel On Her Majesty, James Bond menikahi Tracy Draco. \\
    Namun, di film itu Draco terbunuh. \\
    \hline
  \end{tabular}
  \caption{A sample article (top) and its abstractive summary (bottom).
    Underlined sentences are the extractive summary obtained by following the
    greedy algorithm in~\cite{nallapati2017}.}~\label{fig:sample}
\end{figure*}

\begin{table*}[!t]\fontsize{6.5pt}{6.5pt}\selectfont
  \caption{Corpus statistics.}~\label{tbl:stats}
  \centering
  \begin{tabular}{@{}lrrrrrrrrrrrrrrr@{}}
    \toprule
               & \multicolumn{3}{c}{Fold 1} & \multicolumn{3}{c}{Fold 2} & \multicolumn{3}{c}{Fold 3} & \multicolumn{3}{c}{Fold 4} & \multicolumn{3}{c}{Fold 5} \\
    \cmidrule(lr){2-4} \cmidrule(lr){5-7} \cmidrule(lr){8-10} \cmidrule(lr){11-13} \cmidrule(l){14-16}
                                & train & dev   & test  & train & dev   & test  & train & dev   & test  & train & dev   & test  & train & dev   & test  \\
    \midrule
    \# of articles              & 14262 & 750   & 3762  & 14263 & 749   & 3762  & 14290 & 747   & 3737  & 14272 & 750   & 3752  & 14266 & 747   & 3761  \\
    avg \# of paras / article   & 10.54 & 10.42 & 10.39 & 10.49 & 10.83 & 10.47 & 10.47 & 10.57 & 10.61 & 10.52 & 10.37 & 10.49 & 10.49 & 10.23 & 10.54 \\
    avg \# of sents / para      & 1.75  & 1.74  & 1.75  & 1.75  & 1.75  & 1.75  & 1.75  & 1.74  & 1.73  & 1.74  & 1.73  & 1.77  & 1.75  & 1.79  & 1.74  \\
    avg \# of words / sent      & 18.86 & 19.26 & 18.91 & 18.87 & 18.71 & 19.00 & 18.89 & 18.95 & 18.90 & 18.88 & 19.27 & 18.82 & 18.92 & 18.81 & 18.82 \\
    avg \# of sents / summ      & 3.48  & 3.42  & 3.47  & 3.47  & 3.50  & 3.47  & 3.48  & 3.44  & 3.46  & 3.48  & 3.40  & 3.48  & 3.47  & 3.54  & 3.48  \\
    avg \# of words / summ sent & 19.58 & 19.91 & 19.59 & 19.60 & 19.54 & 19.58 & 19.57 & 19.77 & 19.65 & 19.58 & 19.92 & 19.60 & 19.63 & 19.05 & 19.57 \\
    \bottomrule
  \end{tabular}
\end{table*}

We used a dataset provided by Shortir,\footnote{Formerly http://shortir.com} an
Indonesian news aggregator and summarizer company. The dataset contains roughly
20K news articles. Each article has the title, category, source (e.g., CNN
Indonesia, Kumparan), URL to the original article, and an abstractive summary
which was created manually by a total of 2 native speakers of Indonesian. There
are 6 categories in total: Entertainment, Inspiration, Sport, Showbiz, Headline,
and Tech. A sample article-summary pair is shown in Fig.~\ref{fig:sample}.

Note that 20K articles are actually quite small if we compare to English
CNN/DailyMail dataset used in~\cite{cheng2016} which has 200K articles.
Therefore, we used 5-fold cross-validation to split the dataset into 5 folds of
training, development, and testing set. We preprocessed the dataset by
tokenizing, lowercasing, removing punctuations, and replacing digits with zeros.
We used NLTK~\cite{bird2009} and spaCy\footnote{https://spacy.io} for sentence
and word tokenization respectively.

In our exploratory analysis, we discovered that some articles have a very long
text and some summaries have too many sentences. Articles with a long text are
mostly articles containing a list, e.g., list of songs played in a concert, list
of award nominations, and so on. Since such a list is never included in the
summary, we truncated such articles so that the number of paragraphs are at most
two standard deviations away from the mean.\footnote{We assume the number of
  paragraphs exhibits a Gaussian distribution.} For each fold, the mean and
standard deviation were estimated from the training set. We discarded articles
whose summary is too long since we do not want lengthy summaries anyway.
The cutoff length is defined by the upper limit of the Tukey's boxplot, where for
each fold, the quartiles were estimated from the training set. After removing
such articles, we ended up with roughly 19K articles in total. The complete
statistics of the corpus is shown in Table~\ref{tbl:stats}.

Since the gold summaries provided by Shortir are abstractive, we needed to
label the sentences in the article for training the supervised extractive
summarizers. We followed Nallapati et al.~\cite{nallapati2017} to make
these labeled sentences (called \textit{oracles} hereinafter) using their greedy
algorithm. The idea is to maximize the ROUGE score between the labeled sentences
and the abstractive gold summary. Although the provided gold summaries are
abstractive, in this work we focused on extractive summarization because we
think research on this area are more mature, especially for Indonesian, and thus
starting with extractive summarization is a logical first step toward
standardizing Indonesian text summarization research.

Since there can be many valid summaries for a given article, having only a
single abstractive summary for an article is a limitation of our dataset which we
acknowledge. Nevertheless, we feel that the existence of such dataset is a
crucial step toward a fair benchmark for Indonesian text summarization research.
Therefore, we make the dataset publicly available for others to
use.\footnote{https://github.com/kata-ai/indosum}

\subsection{Evaluation}

For evaluation, we used ROUGE~\cite{lin2004}, a standard metric for text
summarization. We used the implementation provided by
\texttt{pythonrouge}.\footnote{https://github.com/tagucci/pythonrouge}
Following~\cite{cheng2016}, we report the $F_1$ score of R-1, R-2, and R-L.
Intuitively, R-1 and R-2 measure informativeness and R-L measures
fluency~\cite{cheng2016}. We report the $F_1$ score instead of just the recall
score because although we extract a fixed number of sentences as the summary,
the number of words are not limited. So, reporting only recall benefits models
which extract long sentences.

\subsection{Compared methods}

We compared several summarization methods which can be categorized into three
groups: unsupervised, non-neural supervised, and neural supervised methods. For
the unsupervised methods, we tested:
\begin{enumerate}
\item \textsc{SumBasic}, which uses word frequency to rank sentences and selects
  top sentences as the summary~\cite{nenkova2005,vanderwende2007}.
\item \textsc{Lsa}, which uses latent semantic analysis (LSA) to decompose the
  term-by-sentence matrix of a document and extracts sentences based on the
  result. We experimented with the two approaches proposed in~\cite{gong2001}
  and~\cite{steinberger2004} respectively.
\item \textsc{LexRank}, which constructs a graph representation of a document,
  where nodes are sentences and edges represent similarity between two
  sentences, and runs PageRank algorithm on that graph and extracts sentences
  based on the resulting PageRank values~\cite{erkan2004}. In the original
  implementation, sentences shorter than a certain threshold are removed. Our
  implementation does not do this removal to reduce the number of tunable
  hyperparameters. Also, it originally uses \textit{cross-sentence informational
    subsumption} (CSIS) during sentence selection stage but the paper does not
  explain it well. Instead, we used an approximation to CSIS called
  \textit{cross-sentence word overlap} described in~\cite{radev2000} by the same
  authors.
\item \textsc{TextRank}, which is very similar to \textsc{LexRank}
  but computes sentence similarity based on the number of common
  tokens~\cite{mihalcea2004}.
\end{enumerate}
For the non-neural supervised methods, we compared:
\begin{enumerate}
\item \textsc{Bayes}, which represents each sentence as a feature vector and
  uses naive Bayes to classify them~\cite{aone1998}. The original paper
  computes TF-IDF score on multi-word tokens that are identified automatically using
  mutual information. We did not do this identification, so our TF-IDF
  computation operates on word tokens.
\item \textsc{Hmm}, which uses hidden Markov model where states correspond to
  whether the sentence should be extracted~\cite{conroy2001}. The original work
  uses QR decomposition for sentence selection but our implementation does not.
  We simply ranked the sentences by their scores and picked the top 3 as the
  summary.
\item \textsc{MaxEnt}, which represents each sentence as a feature vector and
  leverages maximum entropy model to compute the probability of a sentence
  should be extracted~\cite{osborne2002}. The original approach puts a prior
  distribution over the labels but we put the prior on the weights instead. Our
  implementation still agrees with the original because we employed a bias
  feature which should be able to learn the prior label distribution.
\end{enumerate}

As for the neural supervised method, we evaluated
\textsc{NeuralSum}~\cite{cheng2016} using the original implementation by the
authors.\footnote{https://github.com/cheng6076/NeuralSum} We modified their
implementation slightly to allow for evaluating the model with ROUGE. Note that
all the methods are extractive. Our implementation code for all the methods
above is available online.\footnote{https://github.com/kata-ai/indosum}

As a baseline, we used \textsc{Lead-N} which selects $N$ leading sentences as
the summary. For all methods, we extracted 3 sentences as the summary since it
is the median number of sentences in the gold summaries that we found in our
exploratory analysis.

\subsection{Experiment setup}

Some of these approaches optionally require precomputed term frequency (TF) or
inverse document frequency (IDF) table and a stopword list. We precomputed the
TF and IDF tables from Indonesian Wikipedia dump data and used the stopword list
provided in~\cite{tala2003}. Hyperparameters were tuned to the development
set of each fold, optimizing for R-1 as it correlates best with human
judgment~\cite{lin2003}. For \textsc{NeuralSum}, we tried several scenarios:

\begin{enumerate}
\item tuning the dropout rate while keeping other hyperparameters fixed,
\item increasing the word embedding size from the default 50 to 300,
\item initializing the word embedding with \textsc{FastText} pre-trained
  embedding~\cite{bojanowski2016}.
\end{enumerate}

\noindent Scenario 2 is necessary to determine whether any improvement in
scenario 3 is due to the larger embedding size or the pre-trained embedding. In
scenario 2 and 3, we used the default hyperparameter setting from the authors'
implementation. In addition, for every scenario, we picked the model saved at an
epoch that yields the best R-1 score on the development set.

\section{Results and discussion}

\begin{table*}[!t]
  \caption{Test $F_1$ score of ROUGE-1, ROUGE-2, and ROUGE-L, averaged over 5
    folds.}~\label{tbl:results}
  \centering
  \begin{tabular}{@{}llrrr@{}}
    \toprule
                                           &                                                      & R-1                   & R-2                   & R-L                   \\
    \midrule
                                           & \textsc{Oracle}                                      & 79.27 (0.25)          & 72.52 (0.35)          & 78.82 (0.28)          \\
                                           & \textsc{Lead-3}                                      & 62.86 (0.34)          & 54.50 (0.41)          & 62.10 (0.37)          \\
    \midrule[0.5\lightrulewidth]
    \multirow{4}{*}{Unsupervised}          & \textsc{SumBasic}~\cite{nenkova2005,vanderwende2007} & 35.96 (0.18)          & 20.19 (0.31)          & 33.77 (0.18)          \\
                                           & \textsc{Lsa}~\cite{gong2001,steinberger2004}         & 41.37 (0.19)          & 28.43 (0.25)          & 39.64 (0.19)          \\
                                           & \textsc{LexRank}~\cite{erkan2004}                    & 62.86 (0.35)          & 54.44 (0.44)          & 62.10 (0.37)          \\
                                           & \textsc{TextRank}~\cite{mihalcea2004}                & 42.87 (0.29)          & 29.02 (0.35)          & 41.01 (0.31)          \\
    \midrule[0.5\lightrulewidth]
    \multirow{3}{*}{Non-neural supervised} & \textsc{Bayes}~\cite{aone1998}                       & 62.70 (0.39)          & 54.32 (0.46)          & 61.93 (0.41)          \\
                                           & \textsc{Hmm}~\cite{conroy2001}                       & 17.62 (0.11)          & 4.70 (0.11)           & 15.89 (0.11)          \\
                                           & \textsc{MaxEnt}~\cite{osborne2002}                   & 50.94 (0.42)          & 44.33 (0.50)          & 50.26 (0.44)          \\
    \midrule[0.5\lightrulewidth]
    \multirow{3}{*}{Neural supervised}     & \textsc{NeuralSum}~\cite{cheng2016}                  & 67.60 (1.25)          & 61.16 (1.53)          & 66.86 (1.30)          \\
                                           & \textsc{NeuralSum} 300 emb. size                     & \textbf{67.96 (0.46)} & \textbf{61.65 (0.48)} & \textbf{67.24 (0.47)} \\
                                           & \textsc{NeuralSum} + \textsc{FastText}               & 67.78 (0.69)          & 61.37 (0.93)          & 67.05 (0.72)          \\
    \bottomrule
  \end{tabular}
\end{table*}

\begin{table*}
  \caption{Test $F_1$ score of ROUGE-1 for the out-of-domain
    experiment.}~\label{tbl:ood-results}
  \centering
  \begin{tabular}{@{}llrrrrrr@{}}
    \toprule
                                        & \multirow{2}[2]{*}{Source dom.} & \multicolumn{6}{c}{Target dom.}                                                                     \\
    \cmidrule(l){3-8}
                                        &                                 & Entertainment  & Inspiration    & Sport          & Showbiz        & Headline       & Tech           \\
    \midrule
    \textsc{Oracle}                     &                                 & 75.59          & 81.19          & 77.65          & 78.33          & 80.52          & 80.09          \\
    \textsc{Lead-3}                     &                                 & 51.27          & 52.12          & 67.56          & 65.05          & 65.21          & 50.01          \\
    \textsc{LexRank}                    &                                 & 51.41          & 50.78          & 67.52          & 65.01          & 65.19          & 50.01          \\
    \midrule[0.5\lightrulewidth]
    \multirow{6}{*}{\textsc{NeuralSum}} & Entertainment                   & 52.51          & 53.15          & 72.51          & 67.01          & 67.63          & \textbf{51.81} \\
                                        & Inspiration                     & 52.51          & 52.71          & 72.51          & 67.01          & 68.02          & 51.67          \\
                                        & Sport                           & 52.41          & 53.85          & 72.51          & 66.62          & 68.48          & 50.89          \\
                                        & Showbiz                         & \textbf{53.65} & 49.86          & 72.51          & \textbf{67.81} & 70.88          & 51.22          \\
                                        & Headline                        & 52.80          & \textbf{55.07} & \textbf{72.53} & 67.17          & \textbf{71.59} & 50.92          \\
                                        & Tech                            & 50.39          & 47.93          & 62.43          & 56.93          & 63.44          & 48.00          \\
    \bottomrule
  \end{tabular}
\end{table*}

\subsection{Overall results}

Table~\ref{tbl:results} shows the test $F_1$ score of ROUGE-1, ROUGE-2, and
ROUGE-L of all the tested models described previously. The mean and standard
deviation (bracketed) of the scores are computed over the 5 folds. We put the
score obtained by an oracle summarizer as \textsc{Oracle}. Its summaries are
obtained by using the true labels. This oracle summarizer acts as the upper
bound of an extractive summarizer on our dataset. As we can see, in
general, every scenario of \textsc{NeuralSum} consistently outperforms the other
models significantly. The best scenario is \textsc{NeuralSum} with word
embedding size of 300, although its ROUGE scores are still within one standard
deviation of \textsc{NeuralSum} with the default word embedding size.
\textsc{Lead-3} baseline performs really well and outperforms almost all the
other models, which is not surprising and even consistent with other work that
for news summarization, \textsc{Lead-N} baseline is surprisingly hard to beat.
Slightly lower than \textsc{Lead-3} are \textsc{LexRank} and \textsc{Bayes}, but
their scores are still within one standard deviation of each other so their
performance are on par. This result suggests that a non-neural supervised
summarizer is not better than an unsupervised one, and thus if labeled data are
available, it might be best to opt for a neural summarizer right away. We also
want to note that despite its high ROUGE, every \textsc{NeuralSum} scenario
scores are still considerably lower than \textsc{Oracle}, hinting that it can be
improved further. Moreover, initializing with \textsc{FastText} pre-trained
embedding slightly lowers the scores, although they are still within one
standard deviation. This finding suggests that the effect of \textsc{FastText}
pre-trained embedding is unclear for our case.

\subsection{Out-of-domain results}

Since Indonesian is a low-resource language, collecting in-domain dataset for
any task (including summarization) can be difficult. Therefore, we experimented
with out-of-domain scenario to see if \textsc{NeuralSum} can be used easily for
a new use case for which the dataset is scarce or non-existent. Concretely, we
trained the best \textsc{NeuralSum} (with word embedding size of 300) on
articles belonging to category $c_1$ and evaluated its performance on articles
belonging to category $c_2$ for all categories $c_1$ and $c_2$. As we have a total
of 6 categories, we have 36 domain pairs to experiment on. To reduce
computational cost, we used only the articles from the first fold and did not
tune any hyperparameters. We note that this decision might undermine the
generalizability of conclusions drawn from these out-of-domain experiments.
Nonetheless, we feel that the results can still be a useful guidance for future
work. As comparisons, we also evaluated \textsc{Lead-3}, \textsc{Oracle}, and
the best unsupervised method, \textsc{LexRank}. For \textsc{LexRank}, we used
the best hyperparameter that we found in the previous experiment for the first
fold. We only report the ROUGE-1 scores. Table~\ref{tbl:ood-results} shows the
result of this experiment.

We see that almost all the results outperform the \textsc{Lead-3} baseline,
which means that for out-of-domain cases, \textsc{NeuralSum} can summarize not
just by selecting some leading sentences from the original text. Almost all
\textsc{NeuralSum} results also outperform \textsc{LexRank}, suggesting that
when there is no in-domain training data, training \textsc{NeuralSum} on
out-of-domain data may yield better performance than using an unsupervised model
like \textsc{LexRank}. Looking at the best results, we observe that they all are
the out-of-domain cases. In other words, training on out-of-domain data is
surprisingly better than on in-domain data. For example, for Sport as the target
domain, the best model is trained on Headline as the source domain. In fact,
using Headline as the source domain yields the best result in 3 out of 6 target
domains. We suspect that this phenomenon is because of the similarity between
the corpus of the two domain. Specifically, training on Headline
yields the best result most of the time because news from any domain can be
headlines. Further investigation on this issue might leverage domain similarity
metrics proposed in~\cite{ruder2017}. Next, comparing the best
\textsc{NeuralSum} performance on each target domain to \textsc{Oracle}, we
still see quite a large gap. This gap hints that \textsc{NeuralSum} can still be
improved further, probably by lifting the limitations of our experiment setup
(e.g., tuning the hyperparameters for each domain pair).

\section{Conclusion and future work}

We present \datasetname{}, a new benchmark dataset for Indonesian text
summarization, and evaluated state-of-the-art extractive summarization methods
on the dataset. We tested unsupervised, non-neural supervised, and neural
supervised summarization methods. We used ROUGE as the evaluation metric because
it is the standard intrinsic evaluation metric for text summarization
evaluation. Our results show that neural models outperform non-neural ones and
in absence of in-domain corpus, training on out-of-domain one seems to yield
better performance instead of using an unsupervised summarizer. Also, we found
that the best performing model achieves ROUGE scores that are still
significantly lower than the maximum possible scores, which suggests that the
dataset is sufficiently challenging for future work. The dataset, which
consists of 19K article-summary pairs, is publicly available. We hope that the
dataset and the evaluation results can serve as a benchmark for future research
on Indonesian text summarization.

Future work in this area may focus on improving the summarizer performance by
employing newer neural models such as SummaRuNNer~\cite{nallapati2017} or
incorporating side information~\cite{narayan2017}. Since the gold summaries are
abstractive, abstractive summarization techniques such as attention-based neural
models~\cite{rush2015}, seq2seq models~\cite{nallapati2016}, pointer
networks~\cite{see2017}, or reinforcement learning-based
approach~\cite{paulus2017} can also be interesting directions for future avenue.
Other tasks such as further investigation on the out-of-domain issue, human
evaluation, or even extending the corpus to include more than one summary per
article are worth exploring as well.

% \section*{Acknowledgment}

% more thanks here




\bibliographystyle{IEEEtranBST/IEEEtran}
\bibliography{summ}

\end{document}